\documentclass[a4paper,10pt,twocolumn]{article}
\usepackage[english]{babel}
\usepackage[utf8]{inputenc}
\usepackage[top=1.5cm, left=1.5cm, right=1.5cm, bottom=1.5cm]{geometry}

\renewenvironment{abstract}{\bf\small {\em\ Abstract---}}{}

\usepackage{amsfonts,amssymb,amsmath,amsthm}
\usepackage{subfigure}
\usepackage{graphicx}
\usepackage{array}
\usepackage[footnotesize]{caption}
\usepackage{algorithm}
\usepackage{algorithmic}

\usepackage{amsfonts,amssymb,amsmath,amsthm}

\title{Sparse component separation from Poisson measurements}

\author{I. El Hamzaoui and J. Bobin.\\
  \footnotesize  IRFU, CEA, Universite\'e Paris-Saclay, F-91191 Gif-sur-Yvette, France} 
\date{\empty} % no need for a date

\begin{document}

\maketitle

\begin{abstract} 
Blind source separation (BSS) aims at recovering signals from mixtures. This problem has been extensively studied in cases where the mixtures are contaminated with additive Gaussian noise. However, it is not well suited to describe data that are corrupted with Poisson measurements such as in low photon count optics or in high-energy astronomical imaging ({\it e.g.} observations from the Chandra or Fermi  telescopes). To that purpose, we propose a novel BSS algorithm coined pGMCA that specifically tackles the blind separation of sparse sources from Poisson measurements.
\end{abstract}

\subsection*{Introduction}
\label{sec:introduction}

Multichannel data are composed of $m$ observations ${\bf X}_i$, each of which is made of $t$ samples. According the standard instantaneous linear mixture, each observation is described as a linear combination of $n$ elementary sources or components ${\bf S}_j$. It is classically assumed that the data are corrupted with additive, generally Gaussian, noise, which leads to the following matrix formulation ${\bf X} = {\bf AS} + {\bf N}$, where ${\bf X} \in \mathbb{R}^{m \times t}$ is the observations matrix, ${\bf S} \in \mathbb{R}^{n \times t}$ the sources matrix, ${\bf A} \in \mathbb{R}^{m \times n}$ the mixing matrix and ${\bf N} \in \mathbb{R}^{m \times t}$ for the noise contribution. In this context, BSS aims at recovering both the mixing matrix $\bf A$ and the sources $\bf S$ from the data $\bf X$ only. This is essentially an unsupervised matrix factorization; being ill-posed it requires additional assumptions about the sources and/or mixing matrix such as statistical independence \cite{comon2010handbook}, non-negativity \cite{fevotte2015nonlinear} or sparsity \cite{zibulevsky2001blind,bobin2007sparsity} to only name three. However, the above linear mixture model does not describe precisely data that are commonly found in low photon count imaging, such as in X-ray \cite{badenes2010chandra}. For that purpose, one needs to account for the exact statistics of the measurements, which precisely follow a Poisson distribution. Hence, the data $\bf X$ are only defined statistically from the "pure" mixtures ${\bf AS}$; the probability for a given sample to take the value ${\bf X}_{i}[t]$ is given by the Poisson law: $\mathcal{P} ({\bf X}_{i}[t] \lvert  [{\bf AS}]_i[t]) = \frac{e^{-[{\bf AS}]_i[t]}\, [{\bf AS}]_i[t]^{{\bf X}_{i}[t]}}{{\bf X}_{i}[t]!}$, where $[{\bf AS}]_i[t]$ is the sample of the matrix $\bf AS$ located at the $i$-th row and $t$-th column. To tackle BSS from Poisson measurements, a straightforward approach consists in maximizing the likelihood of mixture variable. In the case of Poisson statistics, this amounts to minimizing the Kullback-Leibler divergence between the data $\bf X$ and the mixture model ${\bf AS}$ with respect to $\bf A$ and $\bf S$, which has been investigated both in the scope of Independent Component Analysis (ICA - \cite{mihoko2002robust}) and Non-negative Matrix Factorization (NMF - \cite{fevotte2015nonlinear}) where it generally refers to robust BSS. In the scope of sparse BSS, the sources are assumed to admit a sparse distribution in some signal representation $\bf \Phi$. While sparse BSS has been successful in various applications \cite{lgmca}, to the best of our knowledge, this blind separation of {\it sparse} sources has not been investigated when the data follow a Poisson distribution. We therefore propose a new algorithm coined poisson Generalized Morphological Component Analysis (pGMCA) to tackle sparse BSS from Poisson measurements.

\subsection*{The pGMCA algorithm}
\label{sec:algo}

In the next, we assume each source admits a sparse representation in some signal representation $\bf \Phi$, which will be assumed to be an orthogonal basis for the sake of simplicity. Next, the sparsity of the sources will be enforced by minimizing their re-weighted $\ell_1$ norm as follows:
\begin{equation}
\label{eq:pBSS}
\min_{{\bf A} \in \mathcal{C} ,{\bf S} \geq 0} \left \| {\bf  \Lambda} \odot {\bf S }{\bf \Phi}^T\right\|_{\ell_1} + \mathcal{L}\left({\bf X} | {\bf A},{\bf S}\right),
\end{equation}
where the matrix $\bf \Lambda$ contains the regularization parameters as well as weigths for re-weighted $\ell_1$ regularization \cite{CandesIRL108}. The second term is the anti-log likelihood of variables $\bf A$ and $\bf S$: $ \mathcal{L}\left({\bf X} | {\bf A},{\bf S}\right) = \sum_{i,t}  [{\bf AS}]_i[t] - {\bf X}_{i}[t] \log\left(  [{\bf AS}]_i[t] \right)$. The convex set $\mathcal{C}$ is the intersection of the positivity constraint ${\bf A} \geq 0$ and the multi-dimensional $\ell_2$-ball constraint that enforces each column of the mixing matrix to have a $\ell_2$ norm lower than $1$ so as to alleviate the scale indeterminacy of the mixture model. The norm $\| \, . \, \|_F$ is the Frobenius norm. Optimizing the problem in Eq. \ref{eq:pBSS} raises several challenges:\\
\textbf{- A multi-convex problem: } the problem is non-convex but convex with respect to each variable $\bf A$ and $\bf S$ assuming the other one is fixed. Hopefully, several optimization strategies have been proposed so far to tackle multi-convex problem such as the Block-Coordinate-Descent algorithm \cite{Tseng01,xu2014globally}, which sequentially optimizes over each variable independently.\\
\textbf{- Non-differentiability and curvature of the data fidelity term: } the data fidelity term $ \mathcal{L}\left({\bf X} | {\bf A},{\bf S}\right)$ is not smooth about $0$, {\it which rigorously excludes the use of BCD}. Furthermore, its curvature at the vicinity of $0$ increases at a hyperbolic rate. Consequently, neglecting the non-differentiability of $\mathcal{L}$ at $0$ to use the BCD would yield to a dramatically slow convergence rate. This would be especially true for measurements that correspond to low flux values.\\
To make use of the BCD, we rather propose to resort to a smooth approximation of $ \mathcal{L}$ based on the smoothing technique introduced by Nesterov in \cite{nesterov2005smooth}. Following Nesterov's approach, a smooth approximate $\mathcal{L}_\mu$ of $\mathcal{L}$ can be built from its Fenchel dual $\mathcal{L}^\star$ \cite{Rockafellar:1970px}:$\mathcal{L}_\mu({\bf Y}) = \mbox{inf}_{\bf U} \, \langle {\bf Y}, {\bf U} \rangle - \mathcal{L}^\star({\bf U}) - \mu \| {\bf U}\|_F^2$. The resulting approximate $\mathcal{L}_\mu({\bf Y})$ is differentiable and its gradient is $\mu$-Lipschitz, which makes the application of BCD possible. The approximate sparse BSS problem to tackle is the following:
\begin{equation}
\label{eq:pGMCA}
\min_{{\bf A} \in \mathcal{C} ,{\bf S} \geq 0} \left \| {\bf  \Lambda} \odot {\bf S }{\bf \Phi}^T\right\|_{\ell_1} + \mathcal{L}_\mu \left({\bf X} | {\bf A},{\bf S} \right).
\end{equation}
Building upon the BCD algorithm, the pGMCA algorithm ({\it e.g.} poisson-GMCA) sequentially updates each variable $\bf S$ and $\bf A$; it is described in Alg.1.

\begin{algorithm}
	\label{algo}
	\caption{pGMCA algorithm}
	\begin{algorithmic} 
		\STATE Initialization ({\it  see below})
		\WHILE{$\left \| {\bf A}^{(k+1)} - {\bf A}^{(k)} \right \|_F > \epsilon$}
		\STATE 1 - {\it Estimating $\bf S$ assuming $\bf A$ is fixed:}
		\STATE \hspace{0.2cm} ${\bf S}^{(k+1)} = \mbox{Argmin}_{\bf S \geq 0}   \left \| {\bf  \Lambda} \odot {\bf S }{\bf \Phi}^T\right\|_{\ell_1} + \mathcal{L}_\mu \left({\bf X} | {\bf A}^{(k)},{\bf S}\right)$
		\STATE 2 - {\it Estimating $\bf A$ assuming $\bf S$ is fixed:}
		\STATE \hspace{0.2cm} ${\bf A}^{(k+1)} = \mbox{Argmin}_{{\bf A} \in \mathcal{C}}  \mathcal{L}_\mu \left({\bf X} | {\bf A},{\bf S}^{(k+1)}\right)$
		\ENDWHILE
	\end{algorithmic}
\end{algorithm}

Step $1$ of the pGMCA algorithm requires minimizing the sum of a smooth data fidelity term with two extra non-smooth penalization, namely the $\ell_1$ norm of the sources in $\bf \Phi$ and the positivity constraint. In the general case, this step does not admit a closed-form expression. Therefore, at each iteration $k$, the estimate ${\bf S}^{(k+1)}$ is obtained numerically using an implementation of the Generalised Forward-Backward Splitting (G-FBS) algorithm. Similarly, step $2$ involves the minimization of $\mathcal{L}_\mu$ subject to ${\bf A} \in \mathcal{C}$, which does not admit a closed-form expression. Since $\mathcal{C}$ is the intersection of a $\ell_2$-ball with unit norm and the positivity orthant, it can be shown that its proximal operator \cite{combettes2011proximal} is explicit. Therefore, step $2$ is solved using a FISTA-like algorithm \cite{BeckFista09}. Practical details of the implementation are described below:\\
{\bf - Initialization: } The problem in Eq. \ref{eq:pGMCA} being multi-convex, the BCD algorithm can be quite sensitive to the initial point. In the next section, the pGMCA algorithm is initialized with the output of GMCA, which generally provides a robust first guess estimate.\\
{\bf - Regularization parameters: }	In practice, the BCD turns out to be quite sensitive to the choice of the regularization parameter. In these preliminary investigations, we did not implement any $\ell_1$ re-weighing scheme; the matrix $\bf \Lambda$ is therefore constant for all the entries of a given source and varies from one source to the other. In practice the regularisation parameter $\lambda_i$ for the $i$-th source is fixed based on the level of the noise that propagates in a single iteration of the G-FBS. More precisely, it is chosen as $\lambda_i = \tau . \mbox{MAD}\left( \nabla_{\bf S} \mathcal{L}_\mu({\bf A}^{(0)},{\bf S}^{(0)}) {\bf \Phi}^T \right)$ where the Median-Absolute-Deviation (MAD) is an empirical estimate of the standard deviation of the noise measured in the gradient in the sparsifying representation. Interestingly, while this is suited for Gaussian noise, such a strategy works quite fine in Poisson case for $\tau = 1$ thanks to the use of a smooth approximation of $\mathcal{L}$. The smoothing parameter mainly depends on the Poisson density $\bf AS$; a small value will yield a more precise approximation but at the cost of a slower convergence ({\it i.e.} the gradient step size in step $1$ and $2$ scales like $\mu$). In the next, it has been set to the mean number of measures counts.\\
{\bf - Stopping criterion and number of iterations: } The number of iterations is fixed $10000$. The algorithm steps when the Frobenius norm between two consecutive estimates of the mixing matrix $\left \| {\bf A}^{(k+1)} - {\bf A}^{(k)} \right \|_F$ is lower than $\epsilon = 10^{-6}$.

\subsection*{Numerical experiments}
\label{sec:numerics}
\begin{figure}[!h]
	\centering
	\includegraphics[width=2.5in]{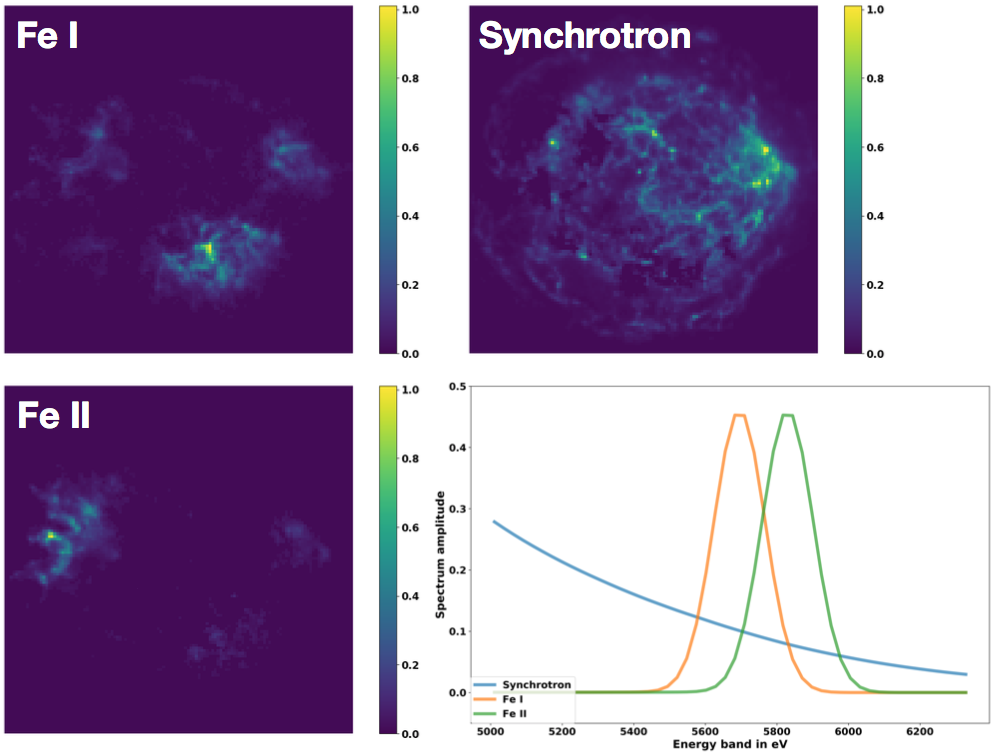}%
	\caption{Input sources and their spectra.}
	\label{fig:input_sources}
\end{figure}

\begin{figure}[!h]
	\centering
	\includegraphics[width=2.5in]{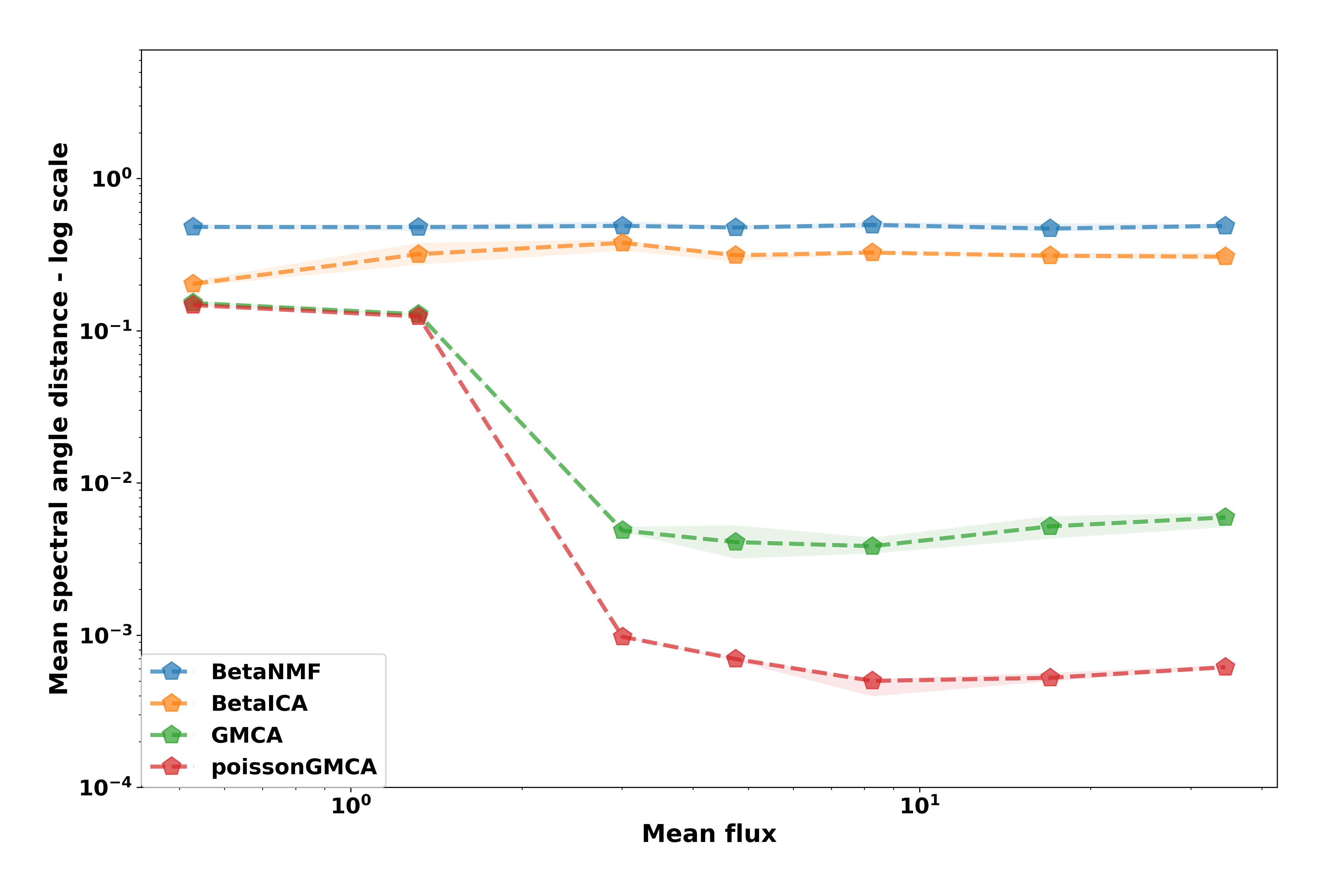}%
	\caption{Spectral angular distance as a function of the mean flux.}
	\label{fig:meanSAD}
\end{figure}

In this section, numerical experiments are carried out on simulations of astrophysical data that have been generated from real Chandra observations of the Cassiopea A supernova remnants. These data are composed of a linear combination of $3$ components ($128 \times 128$ pixels): synchrotron emission, and $2$ redshifted iron emission lines. They are composed of $50$ observations. The electromagnetic spectrum of the synchrotron emission follows a power law while the two iron components have Gaussian-shaped spectra, which are the instrumental response of lines at different redshifted, which are representative of these data in the energy band $5000-6000$ eV (electron-volt). The sources and their spectra are displayed in Fig. \ref{fig:input_sources}.\\
In this abstract, we investigate the performances of the various component separation methods when the mean flux (mean number of counts) evolves. These methods are GMCA, pGMCA, $\beta$-NMF \cite{fevotte2015nonlinear} and $\beta$-ICA \cite{mihoko2002robust}. In the last two ones, the parameter $\beta$ of the $\beta$-divergence is set to $1$ to minimize a Kullback-Leibler divergence, which is well suited for Poisson statistics. $\bf \Phi$ is chosen as isotropic undecimated wavelet \cite{starck2007undecimated}. Since these methods do no impose similar regularization on the sources, they are more fairly compared based on the quality of estimation of the mixing matrix. For that purpose , we make use of the spectral angular distance (SAD) between the estimated $[\hat{\bf A}]^j $ and input $[{\bf A}^o]^j$ columns of the mixing matrix: $\mbox{SAD} = \sum_{j=1}^n \cos^{-1}(| \langle [\hat{\bf A}]^j , [{\bf A}^o]^j \rangle |)/n$. The results are displayed in Fig.\ref{fig:meanSAD}; each point is the mean valued of $10$ Monte-Carlo simulations with different noise realisations. This figure first shows that the robust ICA and robust NMF methods perform badly; the range of mean flux values ({\it i.e.} $0.5$ to $35$) correspond to a rather severe noise level for component separation, which could explain these poor results. For mean flux values larger than $2$ ({\it i.e.} on average a number counts equal to $2$ per pixel), GMCA performs satisfactorily and pGMCA yields a gain of about $1$ order of magnitude with respect to GMCA. For lower number of counts, it is likely that the noise level is too large to perform a reasonable separation process.

\vspace{-0.5cm}

\subsection*{Conclusion}
\label{sec:conclusion}
We investigated the development of a sparsity enforcing method to tackle BSS from measurements with Poisson noise. The proposed pGMCA algorithm builds upon a BCD-like algorithm that makes use a smooth approximation of the Poisson log-likelihood. Preliminary results show a clear improvement of the separation quality with respect to existing methods when the number of counts is low. During the workshop, we will show more extensive results that better highlight the benefits of the proposed pGMCA algorithm. 

\clearpage

%% You can make the bibliography smaller

\end{document}